\documentclass[10pt,twocolumn,letterpaper]{article}

\usepackage{iccv}
\usepackage{times}
\usepackage{epsfig}
\usepackage{graphicx}
\usepackage{amsmath}
\usepackage{amssymb}
\usepackage{multirow}
\usepackage{makecell}

\usepackage{subfigure}
\usepackage{pifont}
\newcommand{\cmark}{$\checkmark$}%
\newcommand{\xmark}{\ding{53}}%

\usepackage[pagebackref=true,breaklinks=true,letterpaper=true,colorlinks,bookmarks=false]{hyperref}

\iccvfinalcopy 

\def\iccvPaperID{6986} 

\newcommand{\add}[1]{\textcolor{black}{#1}}
\newcommand{\edit}[1]{\textcolor{black}{#1}}

\begin{document} 

\title{AD-NeRF: Audio Driven Neural Radiance Fields for Talking Head Synthesis}

\author{\large Yudong Guo\thanks{This work was done when Yudong Guo and Keyu Chen were intern at Dilusense.} \textsuperscript{1,2} \quad Keyu Chen\textsuperscript{1,2} \quad Sen Liang\textsuperscript{3}  \quad Yong-Jin Liu\textsuperscript{4} \quad Hujun Bao\textsuperscript{3} \quad Juyong Zhang\thanks{Corresponding author: juyong@ustc.edu.cn.} \textsuperscript{1} \vspace{1.5 mm}\\
	{\normalsize \textsuperscript{1}University of Science and Technology of China \quad \textsuperscript{2}Beijing Dilusense Technology Corporation} \quad  \\
    \normalsize	\textsuperscript{3}Zhejiang University \quad
	\textsuperscript{4}Tsinghua University  \\
    }
    
\maketitle

\maketitle
\ificcvfinal\thispagestyle{empty}\fi

\begin{abstract}
    Generating high-fidelity talking head video by fitting with the input audio sequence is a challenging problem that receives considerable attentions recently. \add{In this paper, we address this problem with the aid of neural scene representation networks. Our method is completely different from existing methods that rely on intermediate representations like 2D landmarks or 3D face models to bridge the gap between audio input and video output. Specifically, the feature of input audio signal is directly fed into a conditional implicit function to generate a dynamic neural radiance field, from which a high-fidelity talking-head video corresponding to the audio signal is synthesized using volume rendering.} Another advantage of our framework is that not only the head (with hair) region is synthesized as previous methods did, but also the upper body is generated via two individual neural radiance fields. Experimental results demonstrate that our novel framework can (1) produce high-fidelity and natural results, and (2) support free adjustment of audio signals, viewing directions, and background images. Code is available at \href{https://github.com/YudongGuo/AD-NeRF}{https://github.com/YudongGuo/AD-NeRF}.
\end{abstract}

\section{Introduction}
Synthesizing high-fidelity audio-driven facial video sequences is an important and challenging problem in many applications like digital humans, chatting robots, and virtual video conferences. Regarding the talking-head generation process as a cross-modal mapping from audio to visual faces, the synthesized facial images are expected to perform natural speaking styles while synchronizing photo-realistic streaming results as same as the original videos.

Currently, a wide range of approaches have been proposed for this task. Earlier methods built upon professional artist modelling~\cite{edwards2016jali,zhou2018visemenet} or complicated motion capture system~\cite{cao2005expressive,williams2006performance} are limited in high-end areas of the movie and game industry. Recently, many deep-learning-based techniques~\cite{pham2017speech,taylor2017deep,cudeiro2019capture,zhou2019talking,chen2019hierarchical,thies2020neural,wang2021audio2head,zhou2021pose,ji2021audio,zhang2021flow} are proposed to learn the audio-to-face translation by generative adversarial networks (GANs). However, resolving such a problem is highly challenging because it is non-trivial to faithfully relate the audio signals and face deformations, including expressions and lip motions. Therefore, most of these methods utilize some intermediate face representations including reconstructing explicit 3D face shapes~\cite{yi2020audio} and regressing expression coefficients~\cite{thies2020neural} or 2D landmarks~\cite{suwajanakorn2017synthesizing,wang2020mead}. Due to the information loss caused by the intermediate representation, it might lead to semantic mismatches between original audio signals and the learned face deformations. \add{Moreover, existing \edit{audio-driven} methods suffer from several limitations, such as only rendering the mouth part~\cite{suwajanakorn2017synthesizing,thies2020neural} or fixed by static head pose~\cite{pham2017speech,taylor2017deep,cudeiro2019capture,chen2019hierarchical}, thus are not suitable for advanced talking head editing tasks, like pose-manipulation and background-replacement.} 

\add{To address these issues of existing talking head methods, we turn attention to recent developed neural radiance fields (NeRF). We present AD-NeRF, an audio-driven neural radiance fields model that can handle the cross-modal mapping problem without introducing extra intermediate representation.} Different from existing methods which rely on 3D face shape, expression coefficient or 2D landmarks to encode the facial image, we adopt the neural radiance field (NeRF)~\cite{mildenhall2020nerf} to represent the scenes of talking heads. \add{Inspired by dynamic NeRF~\cite{gafni2020dynamic} for modeling appearance and dynamics of a human face, we directly map the corresponding audio features to dynamic neural radiance fields to represent the target dynamic subject.} Thanks to the neural rendering techniques which enable a powerful ray dispatching strategy, our model can well represent some fine-scale facial components like teeth and hair, and achieves better image qualities than existing GAN-based methods. Moreover, the volumetric representation provides a natural way to freely adjust the global deformation of the animated speakers, which can not be achieved by traditional 2D image generation methods. Furthermore, our method takes the head pose and upper body movement into consideration and is capable of producing vivid talking-head results for real-world applications.

Specifically, our method takes a short video sequence, including the video and audio track of a target speaking person as input. Given the audio features extracted via \emph{DeepSpeech}~\cite{amodei2016deep} model and the face parsing maps, we aim to construct an audio-conditional implicit function that stores the neural radiance fields for talking head scene representations. As the movement of the head part is not consistent with that of the upper body part, we further split the neural radiance field representation into two components, one for the foreground face and the other for the foreground torso. In this way, we can generate natural talking-head sequences from collected training data. Please refer to the supplementary video for better visualization of our results.

\begin{figure*}
\begin{center}
  \includegraphics[width=1\linewidth]{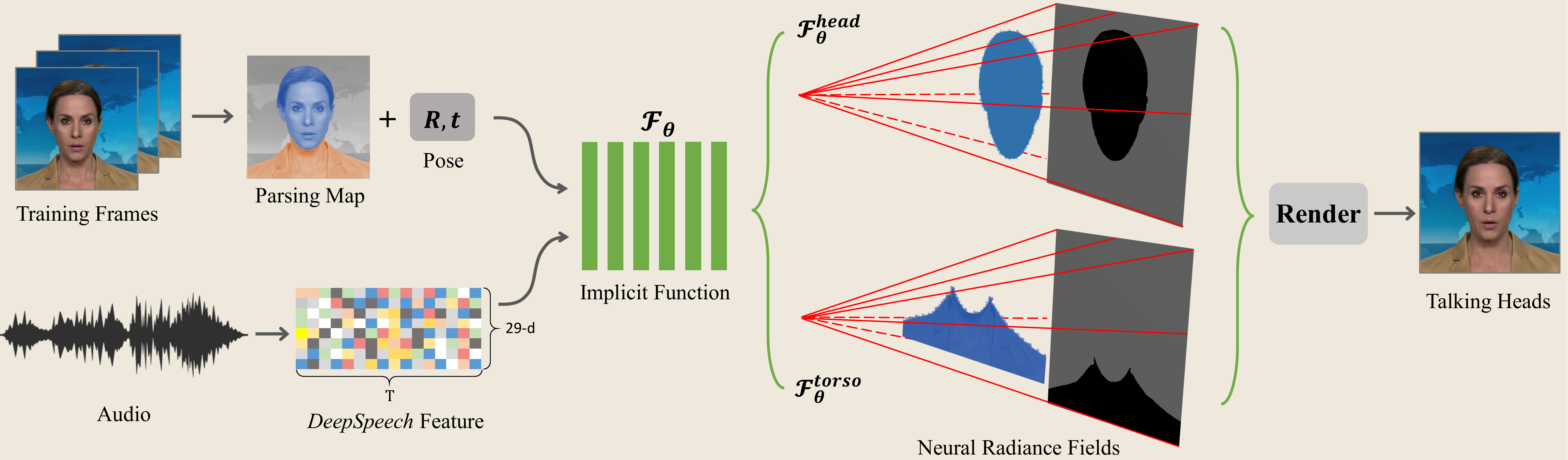}
\end{center}
\vspace*{-4mm}
  \caption{Framework of our proposed talking-head synthesis method. Given a portrait video sequence of a person, we train two neural radiance fields to synthesize high-fidelity talking head with volume rendering.}
\vspace*{-4mm}
\label{fig:framework}

\end{figure*}

In summary, the contributions of our proposed talking-head synthesis method contain three main aspects:
\begin{itemize}
	\vspace*{-2mm}
    \item We present an audio-driven talking head method that directly maps the audio features to dynamic neural radiance fields for portraits rendering, without any intermediate modalities that may cause information loss. Ablation studies show that such direct mapping has better capability in producing accurate lip motion results \edit{with training data of a short video}.
    
    \vspace*{-2mm}
    \item We decompose the neural radiance fields of human portrait scenes into two branches to model the head and torso deformation respectively, which helps to generate more natural talking head results. 
    
    \vspace*{-2mm}
    \item With the help of audio-driven NeRF, our method enables talking head video editings like pose-manipulation and background-replacement, which are valuable for potential virtual reality applications.
\end{itemize}

\section{Related Work}

\noindent {\bf{Audio-driven Facial Animation.}} The goal of audio-driven facial animation is to reenact a specific person in sync with arbitrary input speech sequences. Based on the applied targets and techniques, it can be categorized into two classes: model-based and data-driven methods. The model-based approaches~\cite{schreer2008real,edwards2016jali,zhou2018visemenet} require expertise works to establish the relationships between audio semantics and lip motions, such as phoneme-viseme mapping~\cite{fisher1968confusions}. Therefore, they are inconvenient for general applications except for advanced digital creations like movie and game avatars. With the rise of deep learning techniques, many data-driven methods are proposed to generate photo-realistic talking-head results. Earlier methods try to synthesize the lip motions that fulfill the training data of a still facial image~\cite{bregler1997video,ezzat2002trainable,Chung17b,wiles2018x2face,chen2019hierarchical,vougioukas2019realistic}. Later it is improved to generate full image frames for President Obama by using quantities of his addressing videos~\cite{suwajanakorn2017synthesizing}. Based on the developed 3D face reconstruction~\cite{guo2018cnn,deng2019accurate,Wang2020Lightweight} and generative adversarial networks, more and more approaches are proposed by intermediately estimating 3D face shapes~\cite{karras2017audio,thies2020neural,yi2020audio} or facial landmarks~\cite{zakharov2019few,wang2020mead}. In contrast to our method, they \edit{require more training data due to} the latent modalities, i.e., prior parametric models or low-dimensional landmarks.

\vspace*{-0mm}
\noindent {\bf{Video-driven Facial Animation.}} \edit{Video-driven facial animation is the process of transferring facial pose and expression from a source actor to a target. Most approaches on this task rely on model-based facial performance capture~\cite{thies2015real,thies2016face2face,kim2018deep,kim2019neural}. Thies \etal~\cite{thies2015real} track dynamic 3D faces with RGB-D cameras and then transfer facial expressions from the source actor to the target. Thies \etal~\cite{thies2016face2face} further improve the pipeline by using RGB cameras only. Kim \etal~\cite{kim2018deep} utilize a generative adversarial network to synthesize photo-realistic skin texture that can handle skin deformations conditioned on renderings. Kim \etal~\cite{kim2019neural} analyze the notion of style for facial expressions and show its importance for video-based dubbing.
}

\vspace*{-0mm}
\noindent {\bf{Implicit Neural Scene Networks.}} Neural scene representation is the use of neural networks for representing the shape and appearance of scenes. The neural scene representation networks (SRNs) was first introduced by Sitzmann \etal~\cite{sitzmann2019scene}, in which the
geometry and appearance of an object is represented as a neural network that can be sampled at points in space. Since from last year, Neural Radiance Fields (NeRF)~\cite{mildenhall2020nerf} has gained a lot of attention for neural rendering and neural reconstruction tasks. The underlying implicit representation of the shape and appearance of 3D objects can be transformed into volumetric ray sampling results. Follow-up works extend this idea by using in-the-wild training data including appearance interpolation~\cite{martinbrualla2020nerfw}, introducing deformable neural radiance fields to represent non-rigidly moving objects~\cite{park2020nerfies,pumarola2020d}, and optimizing NeRF without pre-computed camera parameters~\cite{wang2021nerfmm}.

\vspace*{-0mm}
\noindent {\bf{Neural Rendering for Human.}} Neural rendering for human heads and bodies have also attracted many attentions~\cite{fried2019text,liu2020neural,li2021write}. With recent implicit neural scene representations~\cite{saito2019pifu,yang2021stereopifu}, Wang \etal~\cite{wang2020learning} present a compositional 3D scene representation for learning high-quality dynamic neural radiance fields for upper body. Raj \etal~\cite{raj2020pva} adopt pixel-aligned features~\cite{saito2019pifu} in NeRF to generalize to unseen identities at test time. Gao \etal~\cite{Gao-portraitnerf} present a meta-learning framework for estimating neural radiance fields form a single portrait image. Gafni \etal~\cite{gafni2020dynamic} propose dynamic neural radiance fields for modeling the dynamics of a human face. Peng \etal~\cite{peng2020neural} integrate observations across video frames to enable novel view synthesis for human body from a sparse multi-view video.


\vspace*{-2mm}
\section{Method}
\vspace*{-1mm}
\subsection{Overview}
\vspace*{-1mm}
Our talking-head synthesis framework (Fig.~\ref{fig:framework}) is trained on a short video sequence along with the audio track of a target person. Based on the neural rendering idea, we implicitly model the deformed human heads and upper bodies by neural scene representation, i.e., neural radiance fields. In order to bridge the domain gap between audio signals and visual faces, we extract the semantic audio features and learn a conditional implicit function to map the audio features to neural radiance fields (Sec.~\ref{sec:conditionalNeRF}). Finally, visual faces are rendered from the neural radiance fields using volumetric rendering (Sec.~\ref{sec:NeRFRendering}). In the inference stage, we can generate faithful visual features simply from the audio input. Besides, our method can also generate realistic speaking styles of the target person. It is achieved by estimating the neural radiance fields of dynamic heads and upper bodies in a separate manner (Sec.~\ref{sec:two_nerfs}) \edit{with the help of an automatic parsing method~\cite{CelebAMask-HQ} for segmenting the head and torso part and extracting a clean background}. While we transform the volumetric features into a novel canonical space, the heads and other body parts will be rendered differently with their individual implicit models and thus produce very natural results.

\subsection{Neural Radiance Fields for Talking Heads}\label{sec:conditionalNeRF}
Based on the standard neural radiance field scene representation~\cite{mildenhall2020nerf} \edit{and inspired by the dynamic neural radiance fields for facial animation introduced by Gafni \etal~\cite{gafni2020dynamic}}, we present a conditional radiance field of a talking head using a conditional implicit function with an additional audio code as input. Apart from viewing direction $\mathbf{d}$ and 3D location $\mathbf{x}$, the semantic feature of audio $\mathbf{a}$ will be added as another input of the implicit function $\mathcal{F}_{\theta}$. In practice, $\mathcal{F}_{\theta}$ is realized by a multi-layer perceptron (MLP). With all concatenated input vectors $(\mathbf{a}, \mathbf{d}, \mathbf{x})$, the network will estimate color values $\mathbf{c}$ accompanied with densities $\mathbf{\sigma}$ along the dispatched rays. The entire implicit function can be formulated as follows:
\begin{equation}
\mathcal{F}_{\theta}: (\mathbf{a}, \mathbf{d}, \mathbf{x})\longrightarrow (\mathbf{c}, \mathbf{\sigma}).
\end{equation}
\edit{We use the same implicit network structure including positional encoding as NeRF~\cite{mildenhall2020nerf}.}

\noindent\textbf{Semantic Audio Feature.}
In order to extract the semantically meaningful information from acoustic signals, \edit{similar to previous audio-driven methods~\cite{cudeiro2019capture, thies2020neural},} we employ the popular \emph{DeepSpeech}~\cite{amodei2016deep} model to predict a 29-dimensional feature code for each 20ms audio clip. In our implementation, several continuous frames of audio features are jointly sent into a temporal convolutional network to eliminate noisy signals from raw input. Specifically, we use the features $\mathbf{a}\in \mathbb{R}^{16\times 29}$ from the sixteen neighboring frames to represent the current state of audio modality. The usage of audio features instead of regressed expression coefficients~\cite{thies2020neural} or facial landmarks~\cite{wang2020one} is beneficial for alleviating the training cost of intermediate translation network and preventing potential semantic mismatching issue between audio and visual signals.

\begin{figure*}
\begin{center}
  \includegraphics[width=1\linewidth]{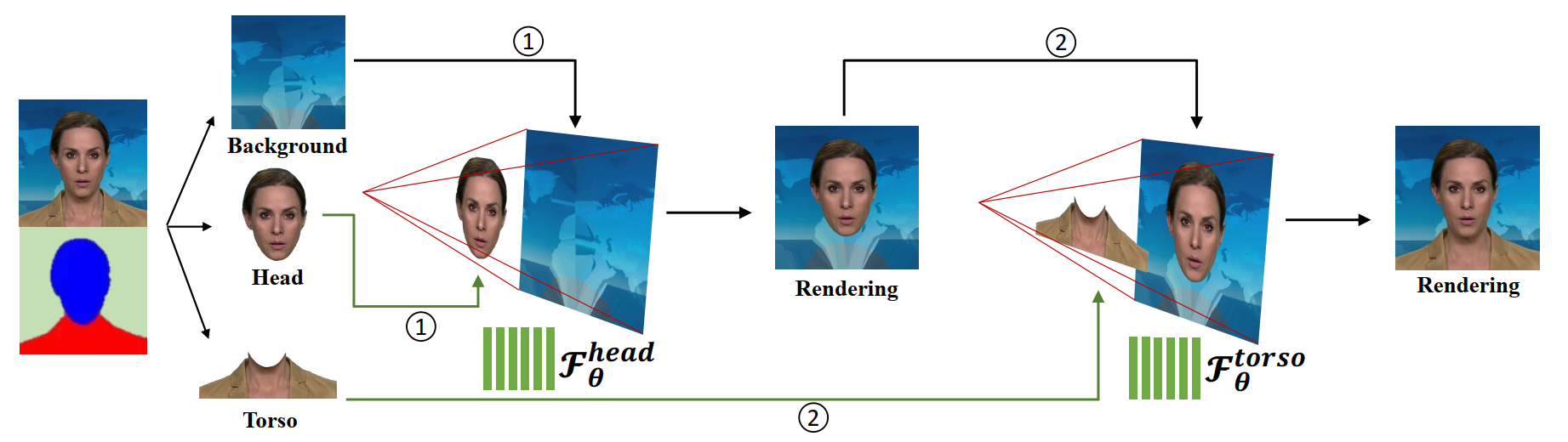}
\end{center}
\vspace*{-5mm}
  \caption{Training process of the two neural radiance fields. We reconstruct the head part and upper-body with Head-NeRF (Step 1) and Torso-NeRF (Step 2) respectively.}
\vspace*{-5mm}
\label{fig:algorithm}
\end{figure*}

\subsection{Volume Rendering with Radiance Fields}\label{sec:NeRFRendering}
With the color $\mathbf{c}$ and density $\mathbf{\sigma}$ predicted by the implicit model $\mathcal{F}_{\theta}$ mentioned above, we can employ the volume rendering process by accumulating the sampled density and RGB values along the rays casted through each pixel to compute the output color for image rendering results. Like NeRF~\cite{mildenhall2020nerf}, the expected color $\mathcal{C}$ of a camera ray $\mathbf{r}(t) = \mathbf{o} + t\mathbf{d}$ with camera center $\mathbf{o}$, viewing direction $\mathbf{d}$ and near bound $t_n$ and far bound $t_f$ is evaluated as:
\begin{equation}~\label{eq:color}
\mathcal{C}(\mathbf{r};\theta,\Pi,\mathbf{a}) = \int_{t_n}^{t_f}\sigma_{\theta}(\mathbf{r}(t)) \cdot \mathbf{c}_{\theta}(\mathbf{r}(t),\mathbf{d}) \cdot T(t)dt,
\end{equation} 
where $\mathbf{c}_{\theta}(\cdot)$ and $\sigma_{\theta}(\cdot)$ are the outputs of the implicit function $\mathcal{F}_{\theta}$ described above. $T(t)$ is the accumulated transmittance along the ray from $t_n$ to $t$:
\begin{equation}
T(t) = exp\left(-\int_{t_n}^{t} \sigma(\mathbf{r}(s))ds\right).
\end{equation} 
$\Pi$ is the estimated rigid pose parameters of the face, represented by a rotation matrix $R\in \mathbb{R}^{3\times 3}$ and a translation vector $t\in \mathbb{R}^{3\times 1}$, i.e., $\Pi=\{R, t\}$. \edit{Similar to Gafni \etal~\cite{gafni2020dynamic}, $\Pi$ is used to transform the sampling points to the canonical space}. Note that during the training stage, we only use the head pose information instead of any 3D face shapes for our network. \edit{We use the two-stage integration strategy introduced by Mildenhall \etal~\cite{mildenhall2020nerf}. Specifically, we first use a coarse network to predict densities along a ray, and then sample more points in areas with high density in the fine network.}

\subsection{Individual NeRFs Representation}\label{sec:two_nerfs}
The reason of taking head pose into account for the rendering process is that, compared to the static background, the human body parts (including head and torso) are dynamically moving from frame to frame. Therefore, it is essential to transform the deformed points from camera space to canonical space for radiance fields training. Gafni \etal~\cite{gafni2020dynamic} try to handle the dynamic movements by decoupling the foreground and background based on the automatic predicted density, i.e., for dispatched rays passing through the foreground pixels, the human parts will be predicted with high densities while the background images will be ignored with low densities. However, there exist some ambiguities to transform the torso region into canonical space. Since the movement of the head part is not consistent with the movement of the torso part and the pose parameters $\Pi$ are estimated for the face shape only, applying the same rigid transformation to both the head and torso region together would result unsatisfactory rendering results in the upper body. To tackle this issue, we model these two parts with two individual neural radiance fields: one for the head part and the other for the torso part.

As illustrated in Fig.~\ref{fig:algorithm}, we initially leverage an automatic face parsing method~\cite{CelebAMask-HQ} to divide the training image into three parts: static background, head and torso. We first train the implicit function for the head part $\mathcal{F}_{\theta}^{head}$. During this step, we regard the head region determined by the parsing map as the foreground and the rest to be background. The head pose $\Pi$ is applied to the sampled points along the ray casted through each pixel. The last sample on the ray is assumed to lie on the background with a fixed color, namely, the color of the pixel corresponding to the ray, from the background image. Then we convert the rendering image of $\mathcal{F}_{\theta}^{head}$ to be the new background and make the torso part to be the foreground. Next we continue to train the second implicit model $\mathcal{F}^{torso}_{\theta}$. In this stage, there are no available pose parameters for the torso region. So we assume all points live in canonical space (i.e., without transforming them with head pose $\Pi$) and add the face pose $\Pi$ to be another input condition (combined with point location $\mathbf{x}$, viewing direction $\mathbf{d}$ and audio feature $\mathbf{a}$) for radiance fields prediction. In other words, we implicitly treat the head pose $\Pi$ as an additional input, instead of using $\Pi$ for explicit transformation within $\mathcal{F}^{torso}_{\theta}$. 

In the inference stage, both the head part model $\mathcal{F}_{\theta}^{head}$ and the torso part model $\mathcal{F}_{\theta}^{torso}$ accept the same input parameters, including the audio conditional code $\mathbf{a}$ and the pose coefficients $\Pi$. The volume rendering process will first go through the head model accumulating the sampled density and RGB values for all pixels. The rendered image is expected to cover the foreground head area on a static background. Then the torso model will fill the missing body part by predicting foreground pixels in the torso region. In general, such an individual neural radiance field representation design is helpful to model the inconsistent head and upper body movements and to produce natural talking head results. 

\vspace*{-5mm}
\add{\subsection{Editing of Talking Head Video}~\label{sec:talking_head_editing}
Since both neural radiance fields take semantic audio feature and pose coefficients as input to control the speaking content and the movement of talking head, our method could enable audio-driven and pose-manipulated talking head video generation by replacing the audio input and adjusting pose coefficients, respectively. Moreover, \edit{similar to Gafni \etal~\cite{gafni2020dynamic}}, since we use the corresponding pixel on the background image as the last sample for each ray, the implicit networks learn to predict low density values for the foreground samples if the ray is passing through a background pixel, and high density values for foreground pixels. In this way, our method decouples foreground-background regions and enables background editing simply by replacing the background image. We further demonstrate all these editing applications in Sec.~\ref{sec:applications}.}

\subsection{Training Details}
\noindent\textbf{Dataset.} For each target person, we collect a short video sequence with audio track for training. The average video length is 3-5 minutes and all in 25 fps. The recording camera and background are both assumed to be static. In testing, our method allows arbitrary audio input such as speech from different identities, gender and language.

\noindent\textbf{Training Data Preprocessing.}
There are three main steps to preprocess the training dataset: (1) We adopt an automatic parsing method~\cite{CelebAMask-HQ} to label the different semantic regions for each frame; (2) We apply the multi-frame optical flow estimation method~\cite{garg2013variational} to get dense correspondences across video frames in near-rigid regions like forehead, ear and hair, and then estimate pose parameters using bundle adjustment~\cite{andrew2001multiple}. It is worth noting that the estimated poses are only effective for the face part but not the other body regions like neck and shoulders, i.e., the face poses could not represent the entire movements of upper body; (3) We construct a clean background image without person (as shown in Fig.~\ref{fig:algorithm}) according to all sequential frames. This is achieved by removing the human region from each frame based on the parsing results and then computing the aggregation results of all the background images. For the missing area, we use Poisson Blending~\cite{perez2003poisson} to fix the pixels with neighbor information.  

\noindent\textbf{Network \& Loss Function.}
In general, our proposed neural radiance field representation network has two main constraints. The first one is the temporal smooth filter. In Sec.~\ref{sec:conditionalNeRF}, we mentioned to process the \textit{DeepSpeech} feature with a window size of 16. The 16 continuous audio features will be sent into a 1D convolutional network to regress the per-frame latent code. In order to assure the stability within audio signals, we adopt the self-attention idea~\cite{thies2020neural} to train a temporal filter on the continuous audio code. The filter is implemented by 1D convolution layers with softmax activation. Hence the final audio condition $a$ is given by the temporally filtered latent code.

Second, we constrain the rendering image of our method to be the same as the training groundtruth. Let $I_r\in \mathbb{R}^{W\times H\times 3}$ be the rendered image and $I_g\in \mathbb{R}^{W\times H\times 3}$ to be the groundtruth, the optmization target is to reduce the photo-metric reconstruction error between $I_r$ and $I_g$. Specifically, the loss function is formulated as:
\begin{equation}
\begin{split}
    &\mathcal{L}_{photo}(\theta) = \sum_{w=0}^{W}\sum_{h=0}^{H}\|I_r(w, h) - I_g(w, h)\|^2,\\
    &I_r(w,h) = \mathcal{C}(r_{w,h};\theta,\Pi,\mathbf{a})
\end{split}
\end{equation}
\section{Experiments}

\subsection{Implementation Details}~\label{sec:implementation}
We implement our framework in PyTorch~\cite{paszke2019pytorch}. Both networks are trained with Adam~\cite{kingma:adam} solver with initial learning rate $0.0005$. We train each model for $400k$ iterations. In each iteration, we randomly sample a batch of $2048$ rays through the image pixels. We train the networks with RTX 3090 and train each model for $400k$ iterations. \edit{For a $5$-minutes video with resolution $450 \times 450$, it takes about 36 hours to train two NeRFs and $12$ seconds to render a frame.}

\begin{figure}[htb]
\begin{center}
  \includegraphics[width=1\linewidth]{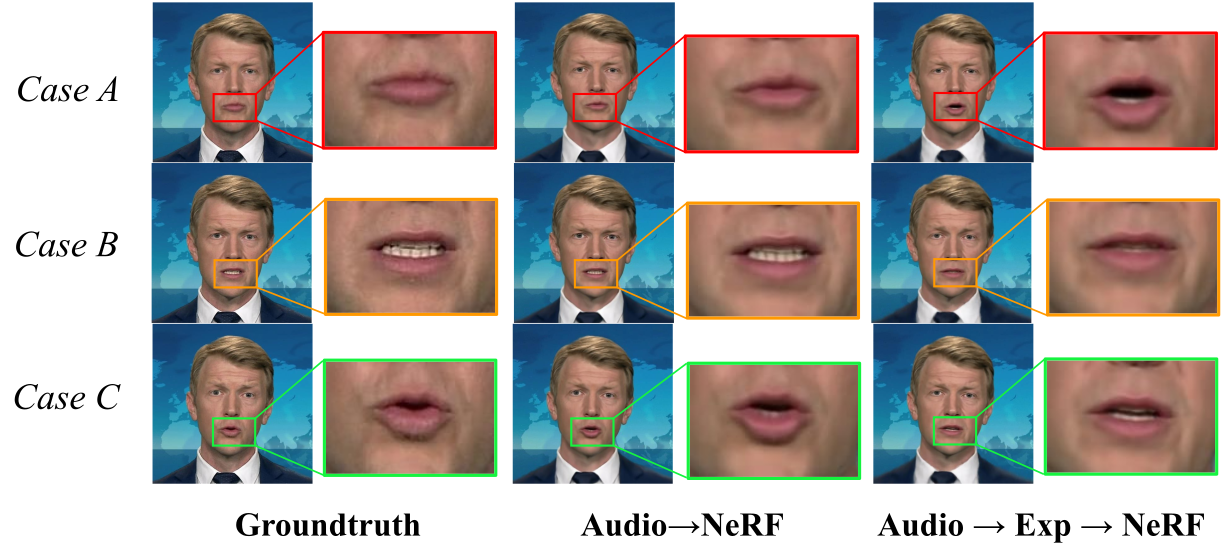}
\end{center}
\vspace*{-4mm}
  \caption{Ablation study on using direct audio or intermediate facial expression representation to condition the NeRF model. It can be observed that direct audio condition has better capability in producing accurate lip motion results.}
\vspace*{-5mm}
\label{fig:ablation_on_audio}
\end{figure}

\begin{figure}[htb]
\begin{center}
  \includegraphics[width=1\linewidth]{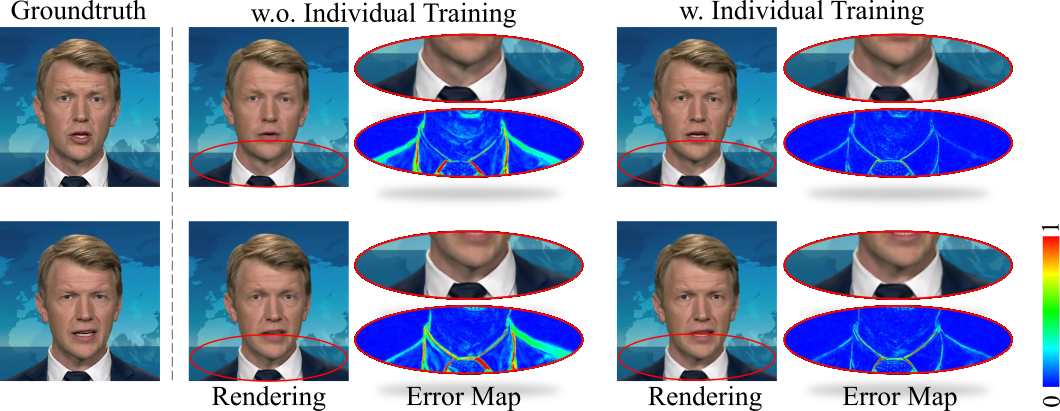}
\end{center}
\vspace*{-3mm}
  \caption{Ablation study on training individual neural radiance field representation for head and torso.}
\vspace*{-5mm}
\label{fig:ablation_on_individual_training}
\end{figure}

\begin{figure*}[htb]
\begin{center}
    \includegraphics[width=1\linewidth]{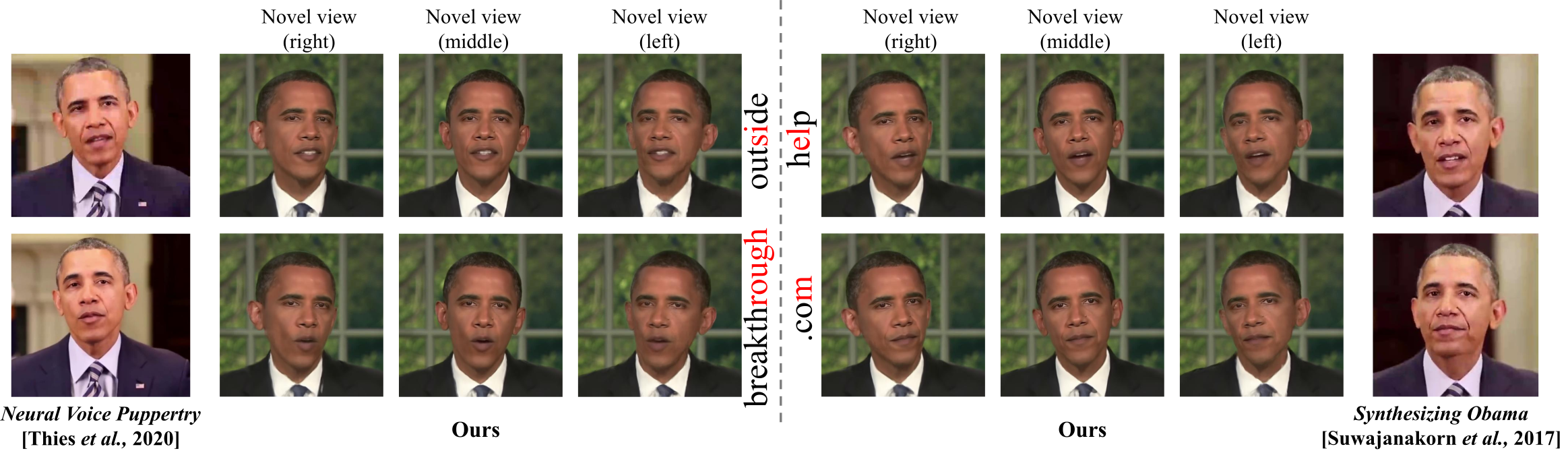}
\end{center}
\vspace*{-4mm}
\caption{Comparison with model-based methods of Thies \etal~\cite{thies2020neural} and Suwajanakorn \etal~\cite{suwajanakorn2017synthesizing}. Our method not only remains the semantics of lip motion, but also supports free adjustment on viewing angles. Please watch our supplementary video for visual results.}
\vspace*{-4mm}
\label{fig:modelbasedcomparison}
\end{figure*}

\begin{figure}[htb]
\begin{center}
  \includegraphics[width=1\linewidth]{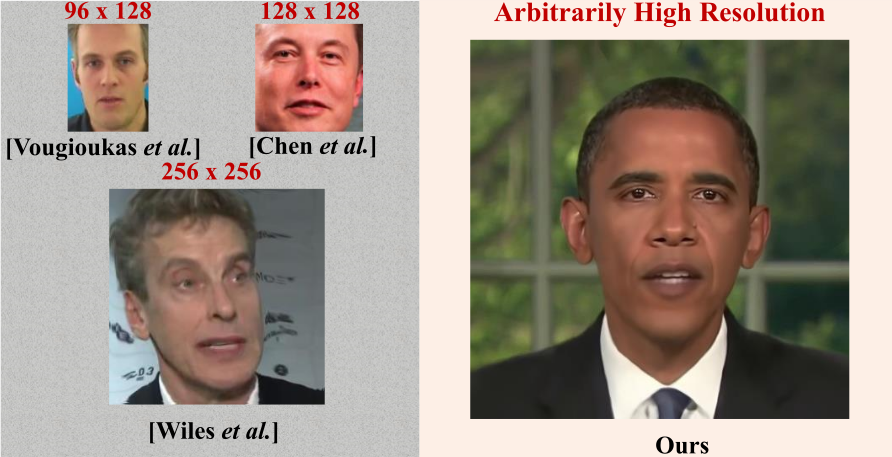}
\end{center}
\vspace*{-4mm}
  \caption{Comparison with image-based methods. The image size decides the image quality of generation results. Please watch our video demo for more results.}
 \vspace*{-4mm}
\label{fig:imagebasedcomparison}
\end{figure}

\begin{figure*}[htb]
\begin{center}
  \includegraphics[width=1\linewidth]{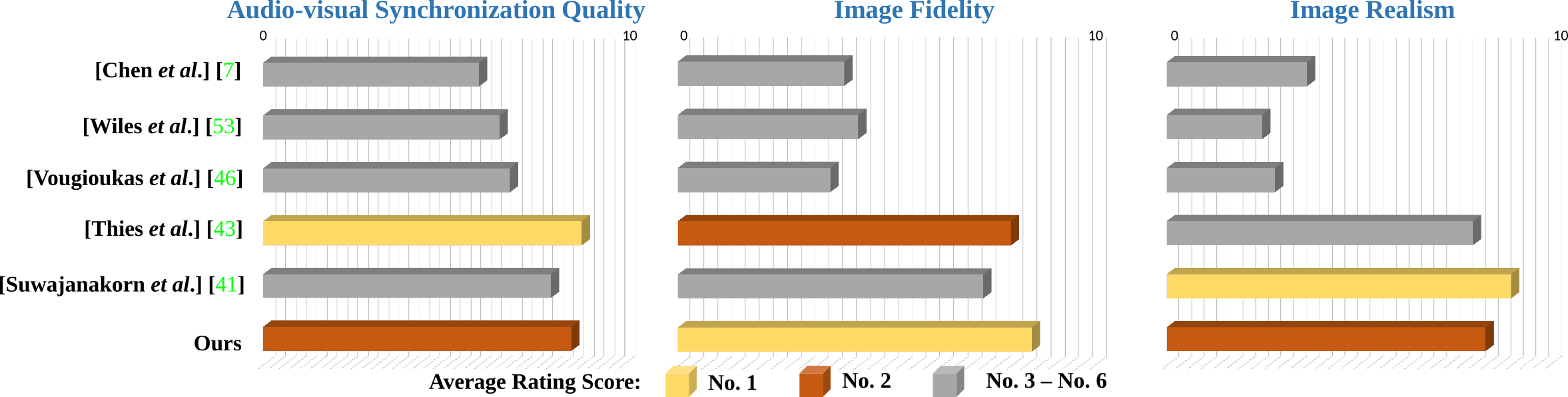}
\end{center}
\vspace*{-4mm}
  \caption{Rating scores from participants. Based on the statics on three different terms, our method achieves comparable results with the other two model-based methods. However, our method only requires a very short video sequence for training, while the other two are trained on multiple and large datasets.}
  \vspace*{-4mm}
\label{fig:userstudy}
\end{figure*}

\begin{figure}[htb]
	\begin{center}
		\includegraphics[width=1\linewidth]{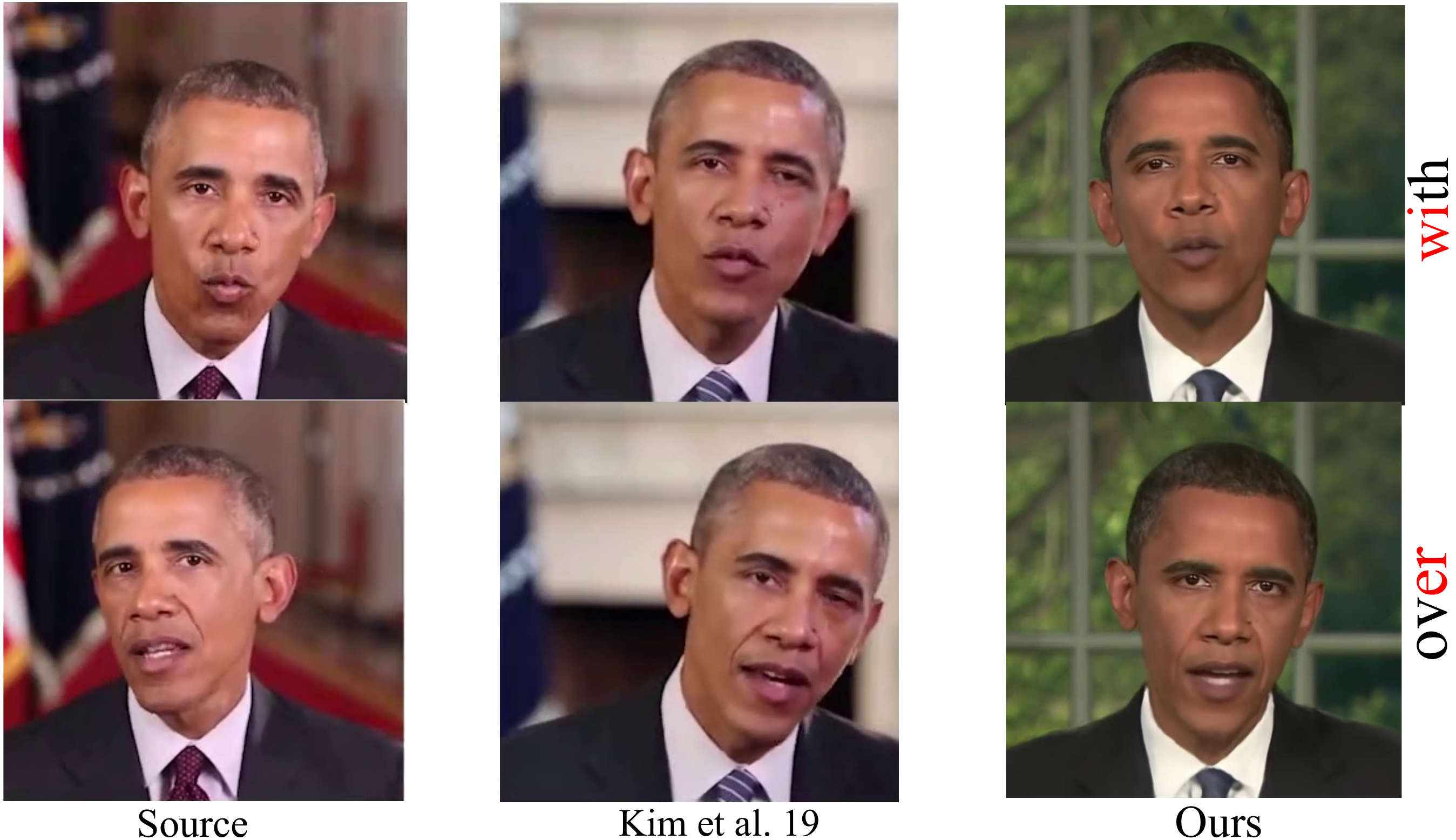}
	\end{center}
\vspace*{-4mm}
	\caption{Comparison with the video-driven method of Kim \etal~\cite{kim2019neural}. On the right are the saying words.}
\vspace*{-4mm}
	\label{fig:compare_video_driven}
\end{figure}

\subsection{Ablation Study}~\label{sec:ablation}
\vspace*{-4mm}

We validate two main components adopted in our framework. First, we compare the neural rendering results based on direct audio condition and additional intermediate condition. Second, we explore the beneficial of training separated neural radiance fields for head and torso region.

\noindent\textbf{Audio condition.} As aforementioned in Sec.~\ref{sec:conditionalNeRF}, our NeRF based talking head model is directly conditioned on audio features to avoid the traininjg cost and information loss within additional intermediate modalities. In Fig.~\ref{fig:ablation_on_audio}, we compare the rendering images generated from audio code and audio-estimated expression code. We use the monocular face tracking method~\cite{thies2016face2face} to optimize expression parameters and use the same network structure as Thies \etal~\cite{thies2020neural} to estimate expression code from audio. From the illustration results, it can be clearly observed that the audio conditioning is helpful for precise lip synchronization.

\noindent\textbf{Individual training for head and torso region.} Another factor we would evaluate is the individual training strategy for head and torso part. To demonstrate the advantages of training two separate neural radiance fields network for these two regions, we conduct an ablative experiment by training a single one NeRF network for the human body movements. In such case, the torso area including neck and shoulders are transformed by the estimated head pose matrices. Therefore there are obviously inaccurate mismatching pixels around the boundary of upper body. We visualize the photo-metric error map of this region for the rendering image and groundtruth. From Fig.~\ref{fig:ablation_on_individual_training}, the illustrated results prove that our individual training strategy is beneficial for better image reconstruction quality.

\edit{We also compute the structural similarity index measure (SSIM) between the generated frames and ground-truth frames on the whole test sequence of 500 frames. The scores are 0.92, 0.88 and 0.87 respectively (higher is better) for our method and the settings of intermediate expression and single NeRF.}

\vspace{0.3in}
\begin{table*}[htb]
\begin{center}
\begin{tabular}{c||cc|cc||c|c|c} \hline
\multirow{2}*{\textbf{Methods}} &\multicolumn{2}{c|}{\textbf{SyncNet score~\cite{Chung16a}$\blacktriangle$}} & \multicolumn{2}{c||}{\textbf{AU error~\cite{OpenFace}$\blacktriangledown$}} &  \multirow{2}*{\textbf{Pose}} & \multirow{2}*{\textbf{Full-frame}} & \multirow{2}*{\textbf{Background}}\\
\multirow{2}*{\textbf{~}} & testset A & testset B & testset A & testset B & & \\ \hline \hline
\lbrack Chen \textit{et al.}\rbrack~\cite{chen2019hierarchical} & 6.129 & 4.388 & 2.588 & 3.475 & \multirow{3}*{static} & \xmark & \xmark \\
\lbrack Wiles \textit{et al.}\rbrack~\cite{wiles2018x2face} & 4.257 & 3.976 & 3.134 & 3.127 &  & \xmark &  \xmark\\ 
\lbrack Vougioukas \textit{et al.}\rbrack~\cite{vougioukas2019realistic} & 5.865 & 6.712 & 2.156 & 2.658 &  & \xmark & \xmark \\ \hline
\lbrack Thies \textit{et al.}\rbrack~\cite{thies2020neural} & 4.932 & - & 1.976 & - & \multirow{2}*{\makecell{copied from \\ source}} & \cmark & \xmark \\ 
\lbrack Suwajanakorn \textit{et al.}\rbrack~\cite{suwajanakorn2017synthesizing} & - & 5.836 & - & 2.176 & & \cmark & \xmark \\ \hline
Ours & 5.239 & 5.411 & 2.133 & 2.287 &  freely adjusted & \cmark & \cmark \\ \hline
Original &5.895 & 6.178 & 0 & 0 & - & - & - \\ \hline
\end{tabular}
\caption{We conduct comparisons on two testsets (A and B) collected from the demos of \textit{Neural Voice Puppertry}~\cite{thies2020neural} and \textit{SynthesizingObama}~\cite{suwajanakorn2017synthesizing}, respectively. $\blacktriangle$ indicates that the confidence value in SyncNet score is better with higher results. $\blacktriangledown$ means that AU error is better with smaller numbers. Moreover, our method can synthesize full-frame imagery while enables pose manipulation and background replacement thanks to the audio-driven neural radiance fields.}
\vspace*{-6mm}
\label{tab:quantitative}
\end{center}
\end{table*}

\vspace*{-7mm}
\subsection{Evaluations}~\label{sec:comparison}
In this section, we compare our method with two categories of talking head synthesis approaches: pure image-based~\cite{wiles2018x2face,chen2019hierarchical,vougioukas2019realistic} and intermediate model-based~\cite{suwajanakorn2017synthesizing,thies2020neural} ones. \add{We conduct both quantitative and qualitative experiments to evaluate the visualized results generated by each method. In the following, we first summarize the compared methods from two categories and then introduce our designed evaluation metrics.}

\begin{figure}[htb]
	\begin{center}
		\includegraphics[width=1\linewidth]{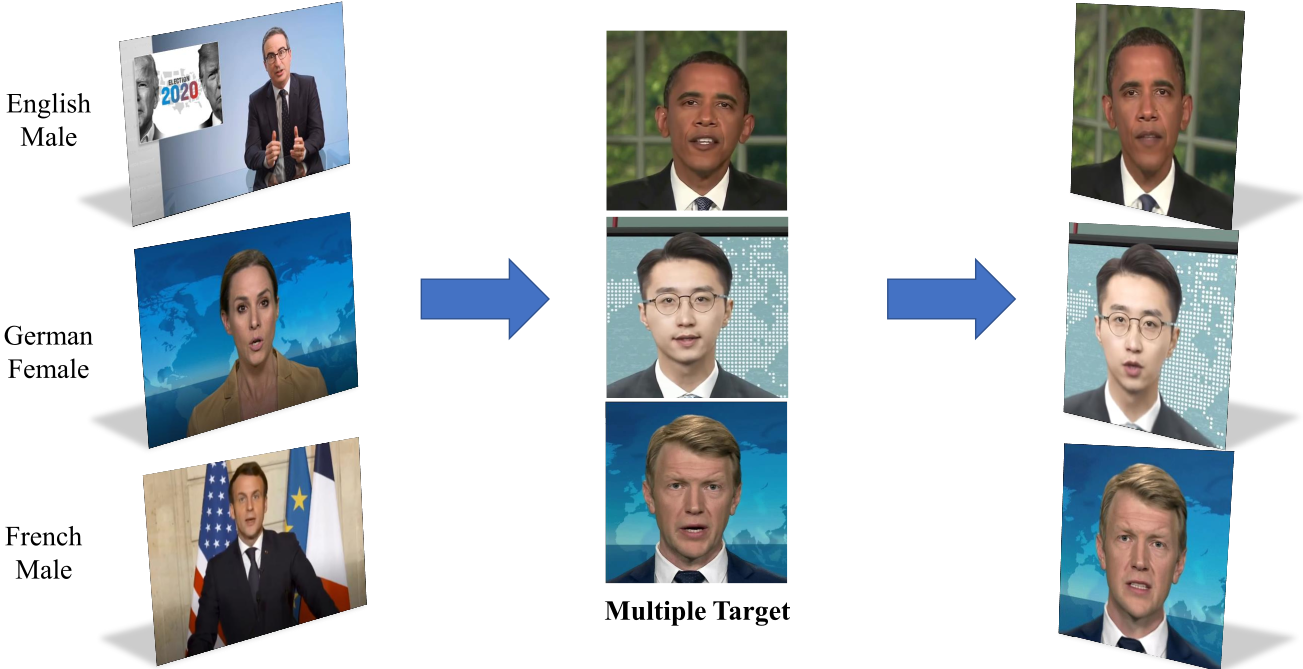}
	\end{center}
	\vspace*{-4mm}
	\caption{Our method allows arbitrary audio input from different identity, gender and language. For the audio-driven results, please refer to our supplementary video.}
	\vspace*{-4mm}
	\label{fig:application_audio}
\end{figure}

\begin{figure*}[htb]
\begin{center}
  \includegraphics[width=1\textwidth]{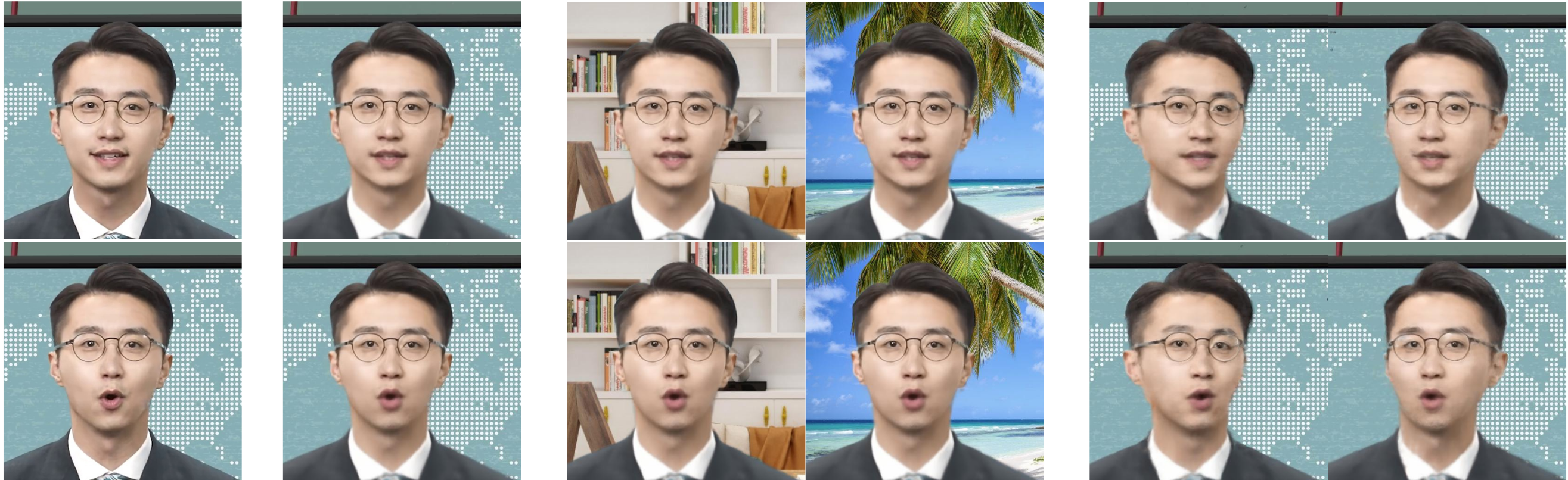}
\end{center}
\vspace*{-4mm}
  \caption{Our method can generate talking head frames with freely adjusted viewing directions and various background images. Each row from left to right: original frames from a video, reconstructed results with audio and pose from the original video, two samples of background-replacement results, two samples of pose-manipulation results.}
\vspace*{-4mm}
\label{fig:application_pose_back}
\end{figure*}

\noindent \textbf{Comparison with Image-based Method.} There are a branch of talking head generation methods~\cite{bregler1997video,ezzat2002trainable,Chung17b,wiles2018x2face,chen2019hierarchical,vougioukas2019realistic} entirely lying in the image domain. Recent deep-learning-based approaches are trained for multiple identities and thus can be applied for new target person. However, the limitation of these methods is obvious as they are only capable of producing still face crop images, and differs from our method that produces full-size images with backgrounds and natural taking styles of target person. In Fig.~\ref{fig:imagebasedcomparison}, we present the audio-driven facial animation results generated by our method and three competitive methods~\cite{wiles2018x2face,chen2019hierarchical,vougioukas2019realistic}. It can be clearly observed that the image-based talking head methods are restricted by the input image size and thus could not producing high-resolution imagery as we do.

\noindent \textbf{Comparison with Model-based Method.} The model-based method refers to the approach that takes prior information in generating photo-realistic face images. The key component of this categorical methods is the statistical model, e.g., PCA model for mouth textures~\cite{suwajanakorn2017synthesizing} or 3D morphable model for face shapes~\cite{thies2020neural}. 

In comparison, we extract the audio from the released demos of the two methods as the input of our framework (we assume the released demos as their best results since both of them did not provide pre-trained model), named as testset A (from \textit{Neural Voice Puppertry}~\cite{thies2020neural}) and testset B (from \textit{SynthesizingObama}~\cite{suwajanakorn2017synthesizing}). In Fig.~\ref{fig:modelbasedcomparison}, we show some selected audio-driven talking head frames from each method. Note that the prior model generally requires large quantities of training data, for example, Suwajanakorn \etal~\cite{suwajanakorn2017synthesizing} reported to use 14 hours high-quality \textit{Obama Addressing} videos for training and Thies \etal~\cite{thies2020neural} took more than 3 hours data for training and 2-3 minutes long video for fine-tuning, while our method only requires a short video clip (3-5 minutes) for training. Despite the huge gap of the training dataset size, our approach is still capable of producing comparable natural results to the other two methods. 

Moreover, our method owns the advantage of freely manipulating the viewing directions on the target person, which means that we can \edit{freely adjust head poses within the range of training data}. We further demonstrate the free-viewing-direction results in Fig.~\ref{fig:application_pose_back} and our supplementary video.

\edit{
\noindent \textbf{Comparison with Video-driven Method.} Besides audio-driven methods, another category of talking head generation methods lie in video-driven, namely driving the target person from a source portrait video. We compare our audio-driven method with a recent style-based video-driven method~\cite{kim2019neural} in Fig.~\ref{fig:compare_video_driven}. We can see that both methods produce high-fidelity talking head results. Note that the method of Kim et al.~\cite{kim2019neural} takes the video frames as input while our method takes the corresponding audio as input. 
} 

\add{
\noindent \textbf{Metrics.} We employ multiple evaluation metrics to demonstrate the superiority of our method to the others. As an audio-driven talking head generation work, the synchronized visual faces are expected to be consistent with audio input while remaining high image fidelity and realistics. To this end, we propose a combined evaluation design, including SyncNet~\cite{Chung16a} scores for audio-visual synchronization quality, Action Units (AU) detection~\cite{AUdetection} (by OpenFace~\cite{OpenFace}) for facial action coding consistency between source and generated results, and a diversified user study on image realism, fidelity and synchronization consistency.} 

\add{
SyncNet~\cite{Chung16a} is commonly used to validate the audio-visual consistency for lip synchronization and facial animation tasks. In this experiment, we use a pretrained SyncNet model to compute the audio-sync offset and confidence of speech-driven facial sequences generated by each comparing method (Tab.~\ref{tab:quantitative}). Higher confidence values are better.}

We employ an action units (AU) detection module by OpenFace~\cite{OpenFace} to compute the facial action units for the source video that providing audio signals and the corresponding generated results. This metric is aimed at evaluating the muscle activation consistency between the source faces and driven ones. The ideal talking-heads are expected to perform similar facial movements as the source faces. We select the lower face and mouth-related AUs as active subjects and compute the mean errors between source and driven faces. The quantitative results are given in Tab.~\ref{tab:quantitative}.

\add{
Finally, we conduct a user study comparisons with the help of 30 attendees. Each participant is asked to rate the talking-head generation results of 100 video clips (9 from Thies \etal~\cite{thies2020neural}, 11 from Suwajanakorn \etal~\cite{suwajanakorn2017synthesizing} and 20 from three image-based methods~\cite{wiles2018x2face,chen2019hierarchical,vougioukas2019realistic} and ours) based on three major aspects: audio-visual synchronization quality, image fidelity and image realism. The head poses for generating results of our method come from a template video clip outside the training set. We collect the rating results within 1 to 10 (the higher the better) and compute the average score that each method gained. The processed statistics are visualized in Fig.~\ref{fig:userstudy}.}
\vspace*{-5mm}

\add{\subsection{Applications on Talking Head Editing}~\label{sec:applications}
As described in Sec.~\ref{sec:talking_head_editing}, our method could enable talking head video editing on audio signal, head movement, and background image. First we show the audio-driven results of the same video with inputs from diverse audio input from different persons in Fig.~\ref{fig:application_audio}. As we can see, our method produces reasonable results with arbitrary audio input from different identities, gender, and language. Then we show the pose-manipulation and background-replacement results of our method in Fig.~\ref{fig:application_pose_back}. We can see that our method allows adjusting viewing directions and various background images replacement for high-fidelity talking portraits synthesis with the trained neural radiance fields. We believe these features would be very exciting for the virtual reality applications like virtual meetings and digital humans.}
\vspace*{-2mm}
\edit{\section{Limitation}
	\vspace*{-2mm}
	We have demonstrated high-fidelity audio-driven talking head synthesis of AD-NeRF. However, our method has limitations. As seen from the supplemental video, for the cross-identity audio-driven results, the synthesized mouth parts sometimes look unnatural due to the inconsistency between the training and driven language. As seen from Fig.~\ref{fig:modelbasedcomparison} and the supplemental video, sometimes the torso parts look blurry due to that the head pose and audio feature cannot totally determine the actual torso movement. 
}

\vspace*{-2mm}
\section{Conclusion}
\vspace*{-2mm}
We have presented a novel method for high-fidelity talking head synthesis based on neural radiance fields. Using volume rendering on two elaborately designed NeRFs, our method is able to directly synthesize human head and upper body from audio signal without relying on intermediate representations. Our trained model allows arbitrary audio input from different identity, gender and language and supports free head pose manipulation, which are highly demanded features in virtual meetings and digital humans.

\small {\noindent{\bf Acknowledgement} This work was supported by the NSFC (62122071, 61725204), the Youth Innovation Promotion Association CAS (No. 2018495) and ``the Fundamental Research Funds for the Central Universities''(No. WK3470000021).}

{\small
\bibliographystyle{ieee_fullname}
\bibliography{egbib}
}

\end{document}